\documentclass[journal]{IEEEtran} 

\IEEEoverridecommandlockouts                              

\usepackage{amsmath,amsfonts}
\usepackage{algorithmic}
\usepackage{algorithm}
\usepackage{array}
\usepackage[caption=false,font=normalsize,labelfont=sf,textfont=sf]{subfig}
\usepackage{textcomp}
\usepackage{stfloats}
\usepackage{url}
\usepackage{verbatim}
\usepackage{graphicx}
\usepackage{cite}
\usepackage{acronym}
\usepackage{xparse} 
\usepackage{amssymb}
\usepackage{algorithmic}
\usepackage{textcomp}
\usepackage{color}
\usepackage{xspace}
\usepackage{acronym}
\usepackage{graphicx}
\usepackage{lettrine}
\usepackage{amssymb}
\usepackage{pifont}%
\usepackage{multirow}
\usepackage{array}
\usepackage{float} 
\usepackage{rotating}
\usepackage{bigstrut}
\usepackage{colortbl}
\usepackage{booktabs}
\usepackage{comment}
\usepackage{stackengine}

\usepackage[vlined,ruled,linesnumbered,algo2e,noend] {algorithm2e}
\usepackage{hyperref}

\SetKwFor{Loop}{loop}{}{end loop}

\newcommand{\mathX}[0]{\ensuremath{X}}

\NewDocumentCommand{\step}{O{\mathX} o}{%
	\IfValueTF{#2}{%
		\ensuremath{\Delta{#1}_{\text{#2}}}%
	}{%
		\ensuremath{\Delta{\mathX}_{\text{#1}}}%
	}%
}

\NewDocumentCommand{\mathJacobian}{m g}{%
	\IfNoValueTF{#2}{%
		\ensuremath{\mathcal{J}_{#1}}
	}{%
		\ensuremath{\mathcal{J}_{#1}(#2)}
	}%
}

\NewDocumentCommand{\mathJacobianTrs}{m g}{%
	\IfNoValueTF{#2}{%
		\ensuremath{\mathtrs{\mathcal{J}}_{#1}}
	}{%
		\ensuremath{\mathtrs{\mathcal{J}}_{#1}(#2)}
	}%
}

\NewDocumentCommand{\mathHessian}{m g}{%
	\IfNoValueTF{#2}{%
		\ensuremath{\text{H}_{#1}}
	}{%
		\ensuremath{\text{H}_{#1}(#2)}
	}%
}
\newcommand{\mathXMAP}[0]{\ensuremath{X^{\text{MAP}}}}

\newcommand{\mathXinitial}[1]{\ensuremath{{#1}^i}}

\newcommand{\mathprn}[1]{\ensuremath{\left(#1\right)}} 
\NewDocumentCommand{\func}{mo}{\ensuremath{#1\IfValueT{#2}{\mathprn{#2}}}}  
\NewDocumentCommand{\norm}{m O{2}}{\ensuremath{\left\lVert#1\right\rVert}^2_{#2}}
\NewDocumentCommand{\normOne}{m O{2}}{\ensuremath{\left\lVert#1\right\rVert}_{#2}}

\newcommand{\matherr}[0]{\ensuremath{{e}}}

\newcommand{\mathinfo}[0]{\ensuremath{{\Omega}}}
\newcommand{\mathtrs}[1]{\ensuremath{{#1}^\text{T}}}
\newcommand{\mathargmax}[1]{\ensuremath{{\underset{#1}{\text{argmax}}}}}
\newcommand{\mathargmin}[1]{\ensuremath{{\underset{#1}{\text{argmin}}}}}
\newcommand{\mathminimize}[1]{\ensuremath{\underset{#1}{\operatorname{min}}}}

\newcommand{\deltemp}{\let\temp\undefined} 

\acrodef{NN}{Neural Network}
\acrodef{RoI}{Region of Interest}
\acrodef{RTK}{Real-Time Kinematic}
\acrodef{SA}{Situational Awareness}
\acrodef{EKF}{Extended Kalman Filter}
\acrodef{DSG}{Dynamic Scene Graph}
\acrodef{UKF}{Unscented Kalman Filter}
\acrodef{VIO}{Visual-Inertial Odometry}
\acrodef{MCL}{Monte Carlo Localization}
\acrodef{MHE}{Moving Horizon Estimation}
\acrodef{GPS}{Global Positioning System}
\acrodef{SURF}{Speeded-Up Robust Features}
\acrodef{CNN}{Convolutional Neural Network}
\acrodef{RNN}{Recurrent Neural Network}
\acrodef{GNN}{Graph Neural Network}
\acrodef{LIDAR}{Light Detection and Ranging}
\acrodef{ATIS}{Asynchronous Time-based Image Sensor}
\acrodef{DAVIS}{Dynamic and Active-pixel Vision Sensor}
\acrodef{AV}{Autonomous Vehicle}
\acrodef{SDF}{Signed Distance Field}
\acrodef{TSDF}{Truncated Signed Distance Field}
\acrodef{ESDF}{Euclidean Signed Distance Field}

\newcommand{\ie}{\textit{i.e., }}
\newcommand{\etal}{\textit{et al. }}

\newcommand\claudiost{\bgroup\markoverwith{\textcolor{violet}{\rule[0.5ex]{2pt}{0.4pt}}}\ULon}
\newcommand\alist{\bgroup\markoverwith{\textcolor{orange}{\rule[0.5ex]{2pt}{0.4pt}}}\ULon}
\newcommand\hridayst{\bgroup\markoverwith{\textcolor{blue}{\rule[0.5ex]{2pt}{0.4pt}}}\ULon}

\newcolumntype{L}[1]{>{\raggedright\arraybackslash\hspace{0pt}}p{#1\linewidth}}
\newcolumntype{M}[1]{>{\centering\arraybackslash\hspace{0pt}}p{#1\linewidth}}
\newcolumntype{R}[1]{>{\raggedleft\arraybackslash\hspace{0pt}}p{#1\linewidth}}

\NewDocumentCommand{\defaulttable}{ O{1.5} O{2pt}  O{10pt} O{.5pt} O{0pt} O{0pt}    }{
\renewcommand{\arraystretch}{#1} 
\setlength{\tabcolsep}{#2} 
\setlength{\extrarowheight}{#3} 
\setlength{\aboverulesep}{#5} 
\setlength{\belowrulesep}{#6} 
}

\definecolor{red}{rgb}{1,0,0} \def\red{\textcolor{red}}
\definecolor{blue}{rgb}{0,0,1} \def\blue{\textcolor{blue}}
\definecolor{forestgreen}{rgb}{0, 0.5,0.5} 
\definecolor{magenta}{rgb}{1,0,1} 
\definecolor{firebrick}{rgb}{0.698039,0.133333,0.133333} 
\definecolor{darkgreen}{rgb}{0,0.392157,0} 
\definecolor{green}{rgb}{0,1,0} 
\definecolor{purple}{rgb}{0.5,0,0.5} 
\definecolor{darkcyan}{rgb}{0,0.545098 ,0.545098} 
\definecolor{goldenrod}{rgb}{0.5,0.5,0}

\usepackage{etoolbox} 

\newbool{reviewmode}

\newcommand{\hide}[2][]{%
  \ifbool{reviewmode}{#1
  }{#2
  }%
}

\boolfalse{reviewmode} 
\title{
Barrier Method for Inequality Constrained Factor Graph Optimization with Application to Model Predictive Control\hide[]{*}
}
\hide[\author{Anonymous Submission for RAL}]{
\author{Anas Abdelkarim$^{1}$, Holger Voos$^{2}$, and Daniel Görges$^{3}$
\thanks{*This research was funded in whole, or in part, by the Luxembourg National Research 
Fund (FNR), MOCCA Project, ref. 17041397. For the purpose of open access, and in fulfilment of the obligations arising from the grant agreement, the author has applied a Creative Commons Attribution 4.0 International (CC BY 4.0) license to any  Author Accepted Manuscript version arising from this submission.}
\thanks{$^{1}$Anas Abdelkarim is with the Interdisciplinary Center for Security, Reliability and Trust (SnT), University of Luxembourg, L-1855 Luxembourg, Luxembourg, and the Department of Electrical and Computer Engineering (EIT), RPTU University Kaiserslautern-Landau, 67663, Germany.  
{\tt\small anas.abdelkarim@uni.lu, abdelkarim@eit.uni-kl.de}}%
\thanks{$^{3}$Holger Voos is with the SnT University of Luxembourg, and the Faculty of Science, Technology, and Medicine (FSTM), Department of Engineering, L-1359 Luxembourg, Luxembourg.  
        {\tt\small holger.voos@uni.lu}}%
\thanks{$^{4}$Daniel Görges is with the EIT, RPTU University Kaiserslautern-Landau    {\tt\small goerges@eit.uni-kl.de}}%
}
}
\begin{document}

\maketitle
\thispagestyle{empty}
\pagestyle{empty}

\begin{abstract}
Factor graphs have demonstrated remarkable efficiency for robotic perception tasks, particularly in localization and mapping applications. However, their application to optimal control problems---especially Model Predictive Control (MPC)---has remained limited due to fundamental challenges in constraint handling. This paper presents a novel integration of the Barrier Interior Point Method (BIPM) with factor graphs, implemented as an open-source extension to the widely adopted g2o framework. Our approach introduces specialized inequality factor nodes that encode logarithmic barrier functions, thereby overcoming the quadratic-form limitations of conventional factor graph formulations.
To the best of our knowledge, this is the first g2o-based implementation capable of efficiently handling both equality and inequality constraints within a unified optimization backend. We validate the method through a multi-objective adaptive cruise control application for autonomous vehicles. Benchmark comparisons with state-of-the-art constraint-handling techniques demonstrate faster convergence and improved computational efficiency.
(Code repository: \hide[\href{https://anonymous.4open.science/r/bipm_g2o}{https://anonymous.4open.science/r/bipm\_g2o})]{\href{https://github.com/snt-arg/bipm\_g2o}{https://github.com/snt-arg/bipm\_g2o})}
\end{abstract}


\section{Introduction}

\IEEEPARstart{F}{actor} graphs have emerged as a powerful paradigm for robotic perception, enabling efficient modeling of large-scale probabilistic inference problems through their sparse graphical structure \cite{dellaert2017factor}. These models are typically solved using gradient-based optimization techniques and have been widely adopted in applications such as Simultaneous Localization and Mapping (SLAM) and situational awareness \cite{bavle2022s}.\\
However, extending factor graphs to optimal control tasks — particularly  Model Predictive Control (MPC) \cite{bemporad2006model,Abdelkarim2020Development} — introduces significant new challenges. Unlike perception problems, control applications require rigorous enforcement of both equality and inequality constraints, e.g. to ensure feasibility and safety \cite{bazzana2024augmented}.\\
While MPC problems are traditionally formulated using domain-specific modeling tools such as CasADi\cite{andersson2019casadi}\hide[,]{ or AMPL \cite{abdelkarim2020optimal,abdelkarim2023optimization},} factor graphs offer a compelling unified framework that bridges perception and control. This integration can enable shared representations across both domains, \cite{abdelkarim2025factor}, simplifying system architecture and enhancing consistency and performance in complex robotic systems.

\hide [The]{Our previous} work \cite{abdelkarim2025ecg2o} extended the factor graph framework to handle equality constraints using a KKT-based Gauss-Newton approach, in which Lagrange multipliers are introduced as additional variable nodes connected to constraint factors. This method embeds constraints directly into the linearized system of equations used for variable updates, preserving both the factor graph’s sparsity and computational efficiency.\\ 
For handling inequality constraints in factor graphs, the Augmented Lagrangian (AL) method~\cite{bertsekas2014constrained} has been the de facto choice due to its compatibility with the quadratic cost structure of standard factor graphs. However, Interior Point Methods (IPMs)~\cite{boyd2004convex,forsgren2002interior} offer superior theoretical properties, including polynomial-time complexity and improved numerical stability, especially for large-scale nonlinear optimization problems. Despite these advantages, IPMs have not yet been systematically integrated into factor graph-based formulations.

The key novelty of this work is the first systematic integration of IPMs into factor graphs via a novel class of inequality factor nodes that encode logarithmic barrier functions. Unlike the quadratic penalties used in AL, these barrier terms are inherently non-quadratic and do not naturally conform to conventional factor graph representations. We address this challenge by introducing specialized inequality factors capable of directly representing logarithmic barriers, thus enabling efficient and direct enforcement of inequality constraints within the factor graph framework.\\
This advancement allows the computational and convergence benefits of IPMs~\cite{bleyer2018advances} to be realized in structured probabilistic optimization settings, while preserving the modularity and interpretability of factor graphs.

We implement our proposed method as an open-source extension to the g2o library and validate it against the state-of-the-art AL-based approach in the factor graph literature. Our implementation preserves a shared front-end abstraction across both BIPM and AL methods, simplifying constraint insertion and emphasizing algorithmic modularity. The resulting new solver is evaluated on a high-performance MPC framework for autonomous driving, specifically for a multi-objective adaptive cruise control (ACC) task. This application is representative of real-world robotics control tasks with multiple critical inequality constraints for safety and comfort.

The complete C++ implementation is publicly available for reproducibility and further development under  \hide[\href{https://anonymous.4open.science/r/bipm_g2o}{https://anonymous.4open.science/r/bipm\_g2o})]{\href{https://github.com/snt-arg/bipm\_g2o}{https://github.com/snt-arg/bipm\_g2o})}.

\section{Related Work}
\noindent  Factor graphs have been extensively studied as an efficient framework for probabilistic inference in robotic perception, particularly in SLAM and situational awareness tasks \cite{dellaert2017factor,grisetti2010tutorial}. The resulting maximum a posteriori (MAP) inference can be formulated as a nonlinear least-squares problem and solved efficiently due to the sparse graphical structure  \cite{dellaert2021factor,kaess2012isam2}. Recent developments, as e.g. reviewed comprehensively in \cite{abdelkarim2025factor}, have extended factor graphs to optimization-based control, also revealing their potential as a unified representation for both perception and control.

A variety of methods have been proposed to incorporate equality constraints into factor graph optimization. A common strategy is to soften constraints by embedding them as weighted penalty terms in the cost function. This technique has been employed in the design of feedback control policies such as Linear Quadratic Regulators (LQRs) using factor graphs  \cite{chen2019LQR}, with applications ranging from trajectory generation in autonomous vehicles \cite{yang2021equality},  to graffiti-drawing robots \cite{chen2022gtgraffiti}, and control of wireless mesh networks \cite{darnley2021flow}. More recently, \hide[related]{our} work proposed a KKT-based Gauss-Newton approach that explicitly incorporates equality constraints into the factor graph via dedicated factors and Lagrange multiplier nodes \cite{abdelkarim2025ecg2o}, enabling efficient constrained optimization while preserving the underlying graph structure.

In contrast, incorporating inequality constraints within factor graphs remains less explored. Xie \etal \cite{xie2020factor} introduced box constraints into a factor graph-based MPC framework using hinge loss functions as soft penalties. Similar approaches were adopted by King-Smith \etal \cite{king2022simultaneous} and Yang \etal \cite{yang2023tightly, yang2024tightly}. However, hinge loss functions penalize constraint violations based on current variable estimates, which are often inaccurate in the early stages of the optimization process. This can result in unnecessary penalties or potentially slowing the convergence.

To address these limitations, Augmented Lagrangian (AL) methods have been explored as a more principled approach for constraint enforcement in factor graph optimization.  The ICS framework by Sodhi \etal \cite{sodhi2020ics} integrates AL with the iSAM algorithm to support incremental constrained smoothing, assuming fixed linearization points. Qadri \etal \cite{qadri2022incopt} extended this framework to handle nonlinear constraints using dynamic relinearization. However, they reported that the algorithm may yield underdetermined systems if an inequality constraint is satisfied and inactivated.  Building on this line of work,  Bazzana \etal \cite{bazzana2024augmented} developed an AL-based solver for SRRG2 and demonstrated its applicability in both pose estimation and real-world MPC tasks. \\
Despite these advances, AL methods are known to be highly sensitive to hyperparameter tuning, including penalty coefficients, their update strategies, and the maximum number of inner and outer iterations. This sensitivity often leads to fragility in practical applications. Our analysis reveals that the tuning parameters used in \cite{bazzana2024augmented} differ significantly from those suitable in our application, highlighting the strong context-dependence of AL-based methods.

In contrast, Barrier Interior Point Methods (BIPMs) offer a compelling alternative. They exhibit faster convergence rates~\cite{steck2018lagrange}, require fewer hyperparameters, and demonstrate greater robustness across a variety of problem settings~\cite{boyd2004convex}. However, BIPMs have not been systematically integrated into factor graph optimization, primarily due to the incompatibility between logarithmic barrier terms and the sum-of-squares cost structure of conventional factor graphs.

This motivates our work, which introduces a novel class of inequality factor nodes that encode barrier terms, enabling the first direct integration of BIPM into factor graph-based control frameworks. This integration fills a major methodological gap and expands the capabilities of factor graphs for constrained optimization in robotics and autonomous systems.
\section{Preliminaries}
\label{sec:Preliminary}
\noindent  Factor graphs are probabilistic graphical models consisting of two types of nodes: variable nodes, which represent unknown states or parameters, and factor nodes, which encode constraints or relationships among these variables. In maximum a posteriori (MAP) inference, the goal is to maximize the joint probability distribution defining the factors, which factorizes according to the graph structure. This can be expressed as
\newcommand{\temp}{\func{\matherr_j}[\mathX]}
\begin{equation}
\begin{aligned}
	\label{eq:MAP_optimization}
 \mathXMAP&= \mathargmax{\mathX}\prod_{j=1}^r \func{\exp}[-\frac{1}{2} \norm{\temp}[\mathinfo_j]]  
 \end{aligned}
\end{equation}
Here, $\mathX$ is the stacked vector containing all variable nodes, $\norm{e}[\mathinfo] = \mathtrs{e} \mathinfo e$ denotes the Mahalanobis norm, $\exp{(.)}$ is the exponential function, and $\mathtrs{.}$ represents the transpose operator for vectors or matrices. The term $\matherr_j$ is the error function associated with factor node $j$, and $\mathinfo_j$ is the positive-definite information matrix of edge $j$, typically the inverse of the covariance matrix.\\ 
The MAP objective can be reformulated as a weighted least squares problem \cite[\S III.B]{abdelkarim2025factor}
\begin{equation}
\label{eq:factor_graph_unconstrainted_optimization}
     \mathXMAP= \mathargmin{\mathX} \quad
        \sum_{j=1}^r   \norm{\temp}[\mathinfo_j], 
 \end{equation}

In robotics perception tasks, the error functions are typically highly nonlinear. Therefore, a first-order linear approximation is applied around a reference point $\mathXinitial{\mathX}$
\begin{equation}
	\begin{aligned}
		\label{eq:linearlization}
		\func{\hat{\matherr}_j}[\mathX] = \matherr_j(\mathXinitial{\mathX}) + \mathJacobian{e_j}{\mathXinitial{\mathX}} [\mathX - \mathXinitial{\mathX}]
	\end{aligned},
\end{equation} 
where $\mathJacobian{e_j}{\mathXinitial{\mathX}}$ is the Jacobian matrix of the error function associated with factor node $j$, evaluated at the linearization point $\mathXinitial{\mathX}$. \\
Applying the first-order optimality condition, which states that the gradient of the cost function vanishes at an optimum, the Gauss-Newton update step can be derived as
\begin{equation}
	\label{eq:sostepg2o}
     \func{\sum_{j=1}^r} 	\underbrace{  \norm{  \mathJacobian{e_j}{{\mathX^i}}}[\mathinfo_j] }_{\text{H}^i_{e_j}}\step[gn]^i =  \sum_{j=1}^r \underbrace{ - \mathtrs{ \mathJacobian{e_j}{{\mathX^i}}} \mathinfo_j \matherr_j ({\mathX}^i)}_{\mathfrak{b}^i_{e_j}}  
\end{equation}
Each factor node contributes a block to the overall linear system used to compute the update step. We refer to such factor nodes as ``cost factor nodes'' to distinguish them from the constraint factor nodes introduced later.

Following~\cite{abdelkarim2025ecg2o}, equality constraints can also be incorporated using factor nodes connected to Lagrange multipliers variables nodes. For an equality constraint $h_j(\mathX) = 0$, the associated factor node is defined by the following error function and information matrix
\begin{equation}
	\label{eq:equality_error}
	\matherr_{h_j} = \left\lbrack \begin{array}{c}
		h_j(\mathX) \\\gamma_{h_j}
	\end{array}\right\rbrack, 
	\quad
	\mathinfo_{h_j} = \left\lbrack \begin{array}{cc} 
		{0}_{d_{h_j}\times d_{h_j}} & {I}_{d_{h_j}\times d_{h_j}} \\
		{I}_{d_{h_j}\times d_{h_j}}  & {0}_{d_{h_j}\times d_{h_j}}
	\end{array}\right\rbrack,
\end{equation}
where both $h_j$ and the Lagrange multipliers $\gamma_{h_j}$ are of dimension $d_{h_j}$, and ${I}$ denotes the identity matrix. The contribution of equality constraints to the linear system of equations is then expressed as
\begin{equation} 
		\text{H}_{h_j}^i  =  \norm{  \mathJacobian{e_{h_j}}{{\mathX^i}}}[\mathinfo_{h_j}], \quad
		\mathfrak{b}^i_{h_j} =  - \mathtrs{ \mathJacobian{e_{h_j}}{{\mathX^i}}} \mathinfo_{h_j} \matherr_{h_j} ({\mathX}^i). 
	\label{eq:equality_edge_contribution}
\end{equation}
In the following section, we describe how inequality constraints are integrated into the factor graph formulation using the Barrier Interior Point Method (BIPM).
\section{Methodology}
\label{sec:Methodology}
\noindent 
This section introduces the BIPM from an optimization standpoint \cite{abdelkarim2023accelerated} and explains how inequality constraints are incorporated into the linear system formulation in factor graphs.
Consider the constrained optimization problem
\begin{subequations}
	\begin{alignat}{5}
		&\mathminimize{\mathX}     &\; \; & \sum_{j=1}^r\norm{\matherr_{j}(\mathX)}[\mathinfo_{j}]\\
		&\text{s.t.} \; &           &h_j(\mathX) =   0&   \; \; \qquad j = 1, \cdots, l 
		\\
		&  &           &g_j(\mathX) \leq   0&   \; \; \qquad  j = 1, \dots, q
	\end{alignat}
\end{subequations}
where $h_j: \mathbb{R}^n \rightarrow \mathbb{R}^{d_{h_j}}$ are equality constraints and $g_j: \mathbb{R}^n \rightarrow \mathbb{R}^{d_{g_j}}$ are inequality constraints.

\subsection{Barrier Interior Point Method} 
The core idea of the Interior Point Method (IPM) is to incorporate inequality constraints into the objective function using a barrier function. With an ideal barrier function  $I(\cdot)$, the optimization problem is reformulated as
\begin{subequations}
	\begin{alignat}{5}
		&\mathminimize{\mathX}     &\; \; & \sum_{j=1}^r\norm{\matherr_{j}(\mathX)}[\mathinfo_{j}] +  \sum_{j=1}^{q} I(g_j(\mathX)), \label{seq:cost_idealBIPM}\\
		&\text{s.t.} \; &           &h_j(\mathX) =   0&   \hspace{-10pt}  j = 1, \dots, l 
	\end{alignat}
	\label{eq:barrier_OP}
\end{subequations}
The ideal barrier function is defined as
\begin{equation}
	\label{eq:Ideal_barrier_function}
	I\left(g\right)=\left\lbrace \begin{array}{cc}
		0 & g\le 0\\
		\infty  & g>0
	\end{array}\right.\;
\end{equation}
The barrier term ($ \sum_{j=1}^{q} I(g_j)$) remains zero as long as the inequality constraints $ g_j \le 0 $ are satisfied, preserving the original objective function.  However, once \mathX \;  exits the feasible region defined by these constraints, the cost function becomes infinite. Thus, to minimize the cost in (\ref{seq:cost_idealBIPM}), the solution \mathX \; must remain strictly within the feasible region.

\begin{figure} [t]
	\centering
	\includegraphics[ width=.48\textwidth]{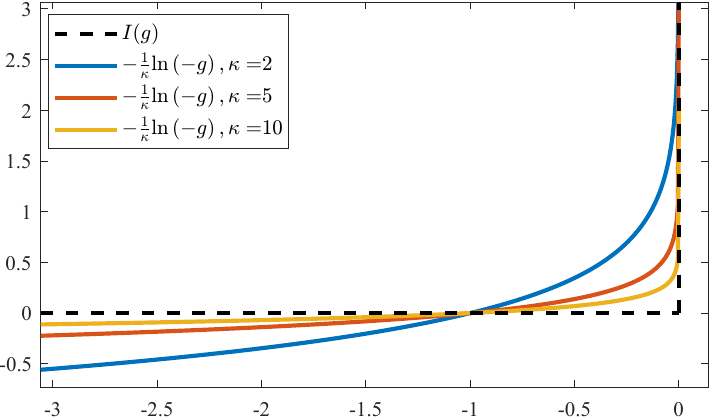}
	\caption{The approximation of the barrier function}
	\label{fig:barrier_fucntion}
\end{figure}

In practice, the ideal barrier function is non-differentiable and therefore unsuitable for gradient-based optimization. To address this, a differentiable logarithmic approximation is used
\begin{equation}
	\label{eq:barrier_function-ipm}
	\hat{I}(g)=- \frac{1}{\kappa} \ln(-g),  \qquad  g<0
\end{equation}
The approximation $\hat{I}(g)$ converges to the ideal barrier  $I(g) $
for $g<0$ as $\kappa > 0$ increases as illustrated in Fig.~\ref{fig:barrier_fucntion}. The resulting approximate optimization problem becomes
\begin{subequations}
	\label{eq:op_approx_barrier}
	\begin{alignat}{5}
		&\mathminimize{\mathX}     &\; \; & \sum_{j=1}^r\norm{\matherr_{j}(\mathX)}[\mathinfo_{j}] -\frac{1}{\kappa} \sum_{j=1}^{q}\ln(-g_j(\mathX))\\
		&\text{s.t.} \; &           &h_j(\mathX) =   0&   \hspace{-20pt}  j = 1, \dots, l 
	\end{alignat}
\end{subequations}

The BIPM (or primal-barrier IPM) solves a sequence of such approximated problems. It starts with a small $\kappa$ and strictly feasible guess-point (\ie satisfying $g_j(\mathX) < 0$). At each iteration, a step is taken using backtracking to ensure the updated solution remains feasible. Once an optimal point is found for the current $\kappa$, the barrier parameter is increased by a factor $ \nu >1 $, and the process repeats until convergence is achieved for a sufficiently large $\kappa$.\\
Although the resulting problem is an equality-constrained optimization, it cannot be directly solved using standard equality-constrained factor graph frameworks. This is because the cost terms derived from the inequality constraints are not quadratic. For instance, attempting to represent the logarithmic barrier term with a factor node defined by
\begin{equation}
	\func{\matherr_{.} }[\mathX] =   \sqrt{-\ln(-g_j(\mathX))}, \text{ and } \mathinfo_{.} = 1/\kappa, 
\end{equation}
is problematic. The argument of the square root is not guaranteed to be non-negative for all feasible \mathX, making the expression potentially undefined or complex. Therefore, special care is required in how such constraints are incorporated into factor graph-based solvers.
\subsection{Inequality Factor Node} 
To incorporate the BIPM into factor graphs, we define specialized inequality factor nodes. These nodes use tailored error functions and information matrices to efficiently encode inequality constraints. For clarity, we present the formulation considering a simplified optimization problem with a single nonlinear inequality constraint $g_j: \mathbb{R}^n \to \mathbb{R}^{d_{g_j}}$. 

Assuming $g_j$ is nonlinear, we linearize it around the current estimate $\mathX^i$, similar to how error functions are handled in conventional cost factor nodes, i.e.
\begin{equation}
	\begin{aligned}
		g_j(\mathX) 
		\approx \func{\hat{g}_j}[\mathX] = { g_ j(\mathXinitial{\mathX}) + \mathJacobian{g_j}{\mathXinitial{\mathX}} [\mathX-\mathXinitial{\mathX}]}
		\label{eq:ineq_linearlization}
	\end{aligned}
\end{equation}
From the BIPM perspective, the optimization problem containing only this inequality factor node becomes
\begin{equation}
	\label{eq:factor_node_OP}
	\mathminimize{\mathX}   \quad  \underbrace{-\frac{1}{\kappa} \sum_{j=1}^{d_{g_j}}\ln(-\func{\hat{g}_{n,j}}[\mathX])}_{f(X)} 		 
\end{equation}
To solve this unconstrained problem, we adopt a Gauss-Newton approach by enforcing the first-order optimality condition, i.e., setting the gradient of $f(\mathX)$ to zero:
\begin{equation}
	\nabla f(\mathX)  =   - \sum_{j=1}^{d_{g_j}}\frac{\overbrace{\nabla \func{\hat{g}_{n,j}}[\mathX]}^{\mathJacobianTrs{g_{n,j}}{\mathXinitial{\mathX}}}}{\kappa \; \func{\hat{g}_{n,j}}[\mathX]}=0
\end{equation}
Using a first-order Taylor expansion around the current estimate, the gradient is approximated as
\begin{equation}
	\label{eq:taylor_expansion}
	\nabla f( \mathXinitial{\mathX} + \Delta \mathX) \approx \nabla f(\mathXinitial{\mathX}) + \mathJacobian{\nabla f}(\mathXinitial{\mathX})   \Delta \mathX = 0, 
\end{equation} 
where  $\mathJacobian{\nabla f}(\mathXinitial{\mathX})$ is the Hessian matrix of $f(\mathX)$ and $\func{\hat{g}_{n,j}}[\mathX^i] = \func{{g}_{n,j}}[\mathX^i]$. This yields the following linear system
\begin{equation}
	\label{eq:GN_update}
  \underbrace{\sum_{j=1}^{d_{g_j}} \frac{\mathJacobianTrs{g_{n,j}}{\mathXinitial{\mathX}} \mathJacobian{g_{n,j}}{\mathXinitial{\mathX}}}{\kappa  [\func{{g}_{n,j}}[\mathX^i]]^2}}_{\text{H}^i_{g_j}}   \Delta \mathX^i =  \underbrace{  \sum_{j=1}^{d_{g_j}}\frac{\mathJacobianTrs{g_{n,j}}{\mathXinitial{\mathX}}}{\kappa \; \func{{g}_{n,j}}[\mathX^i]}}_{\mathfrak{b}^i_{g_j}}
\end{equation}
From this, we define a BIPM-compatible inequality factor node characterized by
\begin{itemize}
	\item an \textit{error function} derived directly from the inequality constraint:
	\begin{equation}
		\matherr_{g_j}(\mathX) =  g_j(\mathX), \\
	\end{equation}	
	\item a \textit{diagonal information matrix} that encodes the barrier scaling:
	\begin{equation}
	 	\mathinfo_{g_j} = \kappa^{-1} \cdot \mathrm{diag}\left([ g_j(\mathX)]^{-2}\right)
	\end{equation}
\end{itemize}
The contribution of the inequality factor to the linear system is then expressed as
 \begin{subequations}
 	\begin{equation}
 		\text{H}_{g_j}^i  =  \norm{  \mathJacobian{e_{g_j}}{{\mathX^i}}}[\mathinfo_{g_j}] 
 	\end{equation}
 	\begin{equation}
 		\mathfrak{b}^i_{g_j} =   \mathtrs{ \mathJacobian{g_j}{{\mathX^i}}} \mathinfo_{g_j} \matherr_{g_j} ({\mathX}^i).
 	\end{equation}
 	\label{eq:ineq_edge_contribution}
 \end{subequations}
Notably, the residual vector $\mathfrak{b}^i_{g_j}$ differs from conventional cost factor terms in that it excludes the negative sign (cf. \eqref{eq:sostepg2o} and \eqref{eq:ineq_edge_contribution}). This distinction is important and must be accounted for when integrating inequality factors into the overall solver framework.

To simplify implementation, the library provides a dedicated class for inequality factor nodes. The factor is defined independently of the underlying solver, while the library automatically constructs the corresponding linear system based on the selected optimization algorithm.

\subsection{BIPM Algorithm for Factor Graph Optimization}
\begin{algorithm2e}[t]
	\SetAlgoNoLine
	\caption{Barrier-IPM Solver for Factor Graphs}
    \label{alg:bipm_solver_factor_graphs}
	\KwIn{%
		\{$ \matherr_j, \mathinfo_j$\}\textsubscript{j=1:r}, 
		\{$ h_j $\}\textsubscript{j=1:l}, 
		\{$ g_j $\}\textsubscript{j=1:q},  	  
		$\kappa > 0$, 
		$\alpha \in (0,1)$,  
		$\epsilon_x > 0$, 
		$\epsilon_g > 0$, 
		$\epsilon_h > 0$, 
		$\kappa_{\text{final}} > 0$, 
		$\nu > 1$, 
		$\gamma_0$, 
		feasible $\mathX$
	}
	\BlankLine  
	\textbf{initialize}  
	$\gamma_{h_j} \gets \gamma_0$ $ \text{for all } j = 1 :l,$ 
	$\mathX \gets [\mathX, \boldsymbol{\gamma}]$,  
	$\text{H} \gets {0}$,  
	$\mathfrak{b} \gets 0$\;
	
	\Repeat{$\kappa  \ge \kappa_{\text{final}}$}{
		\Repeat{$\normOne{\Delta \mathX} < \epsilon_x$  {and} $\normOne{\max(g,0)}[\infty] < \epsilon_g$  and $\normOne{h}[\infty] < \epsilon_h$}{
			\ForEach{$\matherr_j, \mathinfo_j$}{
				$\text{H} \gets \text{H} + \text{H}_{e_j}$\;
				$\mathfrak{b} \gets \mathfrak{b} + \mathfrak{b}_{e_j}$ (Eq.~\ref{eq:sostepg2o})\;
			}
			\ForEach{$h_j$}{
				$\text{H} \gets \text{H} + \text{H}_{e_{h_j}}$\;
				$\mathfrak{b} \gets \mathfrak{b} + \mathfrak{b}_{e_{h_j}}$ (Eq.~\ref{eq:equality_edge_contribution})\;
			}
			\ForEach{$g_j$}{
				$\text{H} \gets \text{H} + \text{H}_{e_{g_j}}$\;
				$\mathfrak{b} \gets \mathfrak{b} + \mathfrak{b}_{e_{g_j}}$ (Eq.~\ref{eq:ineq_edge_contribution})\;
			}
			
			$\Delta \mathX \gets $Solve ($\text{H} \Delta \mathX = \mathfrak{b}$) \\
			Set step size $\zeta \gets 1$\;
			\While{$g(\mathX + \zeta \Delta \mathX) \ge 0$}{
				$\zeta \gets \alpha \zeta$\;
			}
			$\mathX \gets \mathX + \zeta \Delta \mathX$\;
		}
		$\kappa \gets \nu \kappa$\;
	}
	\Return{$\mathX$}	
\end{algorithm2e}
Thus far, we have shown how factor nodes contribute to the construction of a linear system of equations. However, computing update steps is only one part of a complete and robust optimization procedure. While specialized factor node definitions are crucial, additional algorithmic mechanisms are necessary to ensure convergence, feasibility, and numerical stability.

First, we use backtracking line search to maintain feasibility of inequality constraints. Specifically, the step size is iteratively reduced by a factor $0 < \alpha < 1$  until all updated states satisfy  $g(\mathX) \leq 0$. This guarantees that feasibility is preserved throughout the optimization process.
Second, we implement an outer loop to gradually increase the barrier parameter $\kappa$ , thereby refining the approximation of the ideal barrier function and improving solution accuracy. 

Finally, comprehensive termination criteria ensure the optimization process stops only when both optimality and constraint satisfaction have been achieved:
\begin{itemize}
\item Update convergence: $\normOne{\Delta \mathX} < \epsilon_x$
\item Inequality feasibility:  $\normOne{\text{max}(g(\mathX),0)}[\infty] < \epsilon_g$
\item Equality satisfaction: $\normOne{h(\mathX)}[\infty] < \epsilon_h$
\end{itemize}
where $h(\mathX)$, and  $g(\mathX)$ represent the aggregated  equality and inequality constraint vector, respectively.

The complete BIPM solver is summarized in Algorithm~\ref{alg:bipm_solver_factor_graphs}, where vector notation without subscripts denotes the aggregation of all relevant subvectors. The solver maintains feasibility via a backtracking line search and progressively tightens the barrier approximation by increasing the barrier parameter $\kappa$ over iterations.
In addition, the dimensions of the system matrix $\text{H}$ and the residual vector $\mathfrak{b}$ correspond to the full set of optimization variables, including the Lagrange multipliers $\gamma$. All Jacobians are computed with respect to this set of variables (including the Lagrange multipliers).
In practical implementations, each factor contributes local blocks of $\text{H}$ and $\mathfrak{b}$ using only the variables it is connected to. These local contributions are then mapped into the global system using sparse indexing, ensuring memory efficiency and scalability. Nevertheless, the algorithm is presented in terms of global matrix operations for conceptual clarity. Moreover,  the practical implementation includes a maximum number of total iterations, as well as a limit on the number of iterations for the inner loop.

 \section{Results and Performance Evaluation}
\noindent  This section presents a comparative evaluation of the proposed BIPM-based solver against an Augmented Lagrangian (AL) baseline, applied to the Multi-objective Adaptive Cruise Control (MACC) problem for battery electric vehicles (BEVs) formulated as an MPC task.

\subsection{Factor Graph Representation of MACC} 
\noindent 
Following the framework in~\cite{jia2023performance}, we reformulate the MACC problem from the perspective of factor graphs and present its graphical structure. The optimization problem for MACC is given by
\begin{subequations}	
	\begin{equation}
		\begin{aligned}
			\mathminimize{\mathX} \sum_{i=k}^{k+N-1} &(
			\underbrace{\norm{p(v_i,F_{t,i})}[\mathinfo_p]}_{\phi_1} + 
			\underbrace{\norm{F_{b,i}}[w_1]}_{\phi_2} + 
			\underbrace{\norm{\delta_{{\text{far}},i}}[w_2]}_{\phi_3}\\[8pt]
			&+ 
			\underbrace{\norm{\delta_{{F_t},i}}[w_3]}_{\phi_4} + 
			\underbrace{\norm{\delta_{{F_b},i}}[w_4]}_{\phi_5})
		\end{aligned}
	\end{equation} 
	\vspace*{-1.5\baselineskip} 
	\begin{align}
		\text{subject to}	\hspace*{5.1cm} \nonumber	&  \\
		v_{i+1} - \left(v_{i} + \frac{T_s}{m_{\text{eq}}}(F_{t,i} - F_{b,i} - \bar{F}_{\text{resist},i})\right) &= 0  \label{subeq:MACC_b}\\
		d_{i+1} - d_i - \frac{T_s}{2}\left({v_{p,i} + v_{p,i+1}} - {v_{i} + v_{i+1}}\right) &= 0\label{subeq:MACC_c}\\         
		d_{\min} + h_{\text{safety}} v_{i+1} - d_{i+1} &\leq 0 \label{subeq:MACC_d}\\ 
		d_{i+1} - \left(d_{\min} + h_{\text{track}} v_{i+1}\right) + \delta_{\text{far},i} &\leq 0 \label{subeq:MACC_e} \\	
		-F_{t,i} &\leq 0 \label{subeq:MACC_f}\\
		F_{t,i} - F_{t,\max} &\leq 0\label{subeq:MACC_g}\\ 
		F_{t,i} -\left(a_{20} + a_{21} v_{i}\right) &\leq 0 \label{subeq:MACC_h}\\	
		-F_{b,i} &\leq 0 \label{subeq:MACC_i}\\
		F_{b,i} - F_{b,\max} &\leq 0 \label{subeq:MACC_j}\\
		v_{i+1} - v_{\max,i+1} &\leq 0 \label{subeq:MACC_k}\\
		-v_{i+1} &\leq 0 \label{subeq:MACC_l}\\
		|F_{t,i} - F_{t,i-1}| -\delta_{{F_t},i} &\leq 0 \label{subeq:MACC_m}\\
		|F_{b,i} - F_{b,i-1}| -  \delta_{{F_b},i} &\leq 0 \label{subeq:MACC_n}
	\end{align}
\end{subequations} 
Here, $\mathX = \mathtrs{[v,d,F_t,F_b,\delta_{\text{far}}, \delta_{{F_t}},\delta_{{F_b}}]}$ represents the variable nodes of the factor graph: host vehicle velocity, inter-vehicle distance, traction and braking forces, and penalty slack variables. $T_s$ denotes the sampling time over the horizon, and $i \in \{k, \dots, k+N-1\}$ indexes the control horizon. The term $v_p$ is the velocity of the preceding vehicle, and $\bar{F}_{\text{resist}}$ is a linear model of resistive forces including gravity, rolling resistance, and aerodynamic drag. Moreover,  \( h_{\text{safety}} \) and \( h_{\text{track}} \) are constant parameters that reflect time headways used to define safe and tracking inter-vehicle distances, respectively. \\
The cost function encompasses multiple objectives: $\phi_1$ promotes energy efficiency, $\phi_2$  penalizes unnecessary braking, which causes a dissipation of the vehicle’s kinetic energy, $\phi_3$ promotes tracking the preceding vehicle, and $\phi_{4/5}$ address ride comfort by smoothing force variations. The cost terms $\phi_2$ to $\phi_5$ naturally map onto factor nodes in the graphical representation. However, 
the  power consumption term  $\phi_1$ in the original formulation is presented by a convex quadratic expression
\begin{equation}
	\begin{aligned}
		P(v_{i}, F_{t,i}) = w_1 T_s \Bigg( 
		& \mathtrs{\left[
			\begin{array}{c}
				v_{i} \\
				F_{t,i}
			\end{array}
			\right]}
		\underbrace{
			\left[
			\begin{array}{cc}
				2 p_{20} &  p_{11} \\
				p_{11} & 2 p_{02}
			\end{array}
			\right]
		}_{Q_c}
		\left[
		\begin{array}{c}
			v_{i} \\
			F_{t,i}
		\end{array}
		\right] \\
		& + 
		\mathtrs{\underbrace{
				\left[
				\begin{array}{c}
					p_{10} \\
					p_{01}
				\end{array}
				\right]
			}_{b_c}}
		\left[
		\begin{array}{c}
			v_{i} \\
			F_{t,i}
		\end{array}
		\right]
		+ p_{00}
		\Bigg)
	\end{aligned}
\end{equation}
This can be rewritten in canonical quadratic form by completing the square, which is valid since \(Q_c\) is symmetric and positive definite — a common property in MPC formulations. The cost term thus becomes
\begin{equation}
	\phi_1 = 
	\norm{
		\left[
		\begin{array}{c}
			v_{i} \\
			F_{t,i}
		\end{array}
		\right]
		+ \tfrac{1}{2} Q_c^{-1} b_c
	}[{w_1 T_s Q_c}] 
	+ \text{const},
\end{equation}
where the constant term can be safely omitted from the optimization objective, as it does not affect the optimal solution. 

The dynamics of the host vehicle's velocity and the inter-vehicle distance are governed by constraints~\eqref{subeq:MACC_b} and~\eqref{subeq:MACC_c}, respectively. The safety constraint~\eqref{subeq:MACC_d} ensures that a minimum safe following distance is preserved, while~\eqref{subeq:MACC_e} encourages the vehicle to actively track its preceding vehicle.

Constraints~\eqref{subeq:MACC_f} to~\eqref{subeq:MACC_j} limit the traction and braking forces based on the physical capabilities of the motor and brake systems. The parameters $a_{20}$ and $a_{21}$ are model coefficients accounting for the vehicle's driveline characteristics.

Speed limits are enforced through~\eqref{subeq:MACC_k} and~\eqref{subeq:MACC_l}, ensuring compliance with traffic laws and road conditions. Finally, the rate constraints~\eqref{subeq:MACC_m} and~\eqref{subeq:MACC_n} regulate the variation in traction and braking forces to promote passenger comfort by avoiding aggressive control actions.

\begin{figure}[!htbp]
	\centering
 \includegraphics[trim={0cm .6cm 0cm .3cm}, width=1.05\linewidth]{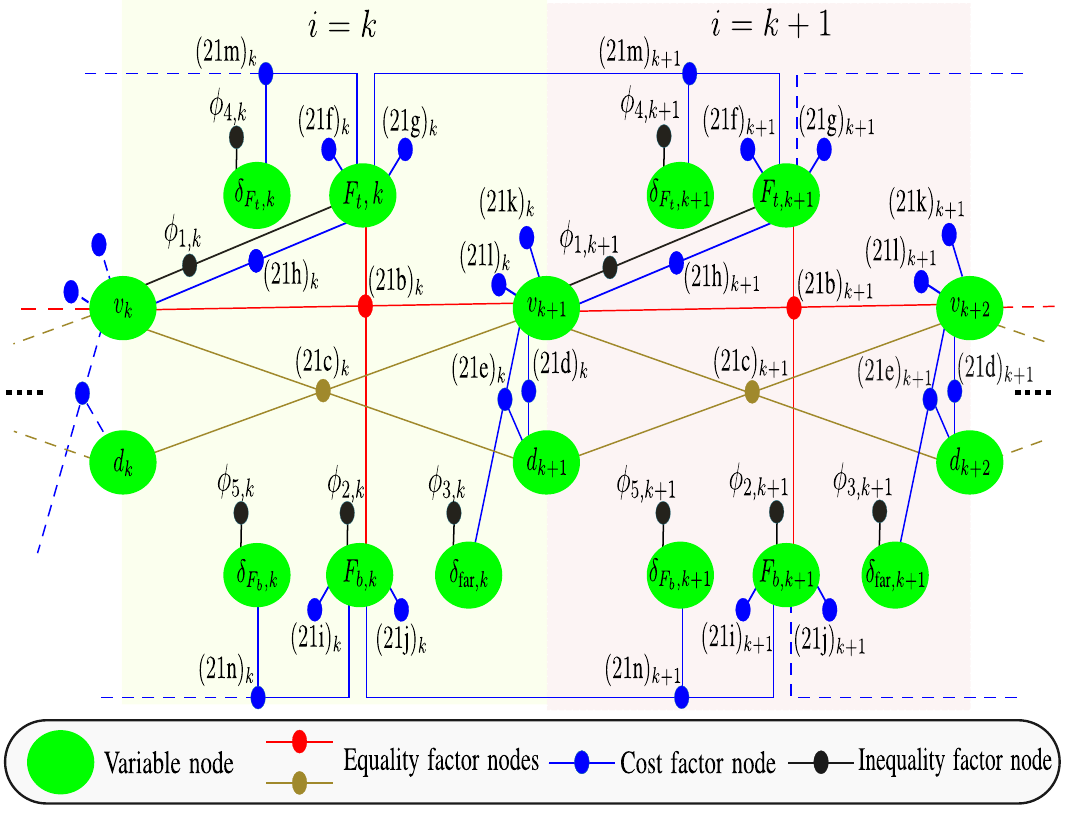}
 \caption{Factor graph representation of the  MACC problem}   
 \label{fig:MACC_FG}
\end{figure}

Figure \ref{fig:MACC_FG} illustrates the factor graph formulation for the MACC task. Despite the variety of factor nodes (equality, inequality, and cost constraints), the graph retains a sparse structure, as each node connects only to a subset of others within the same time step. Dashed edges indicate optional temporal dependencies, active only when adjacent time steps (e.g., $k -1$  or $k + 2$) are considered. For visual clarity, equality constraints are highlighted with distinct edge colors.

\subsection{Experimental Setup}
\noindent 
The experimental evaluation compares two optimization approaches: the BIPM and the AL method. Both solvers were implemented on identical hardware platforms, consisting of an Intel Core i9 12th Gen processor and 32 GB of RAM running Ubuntu 22.04. To ensure a fair comparison, both methods used the same MPC parameters, constraint settings, and stopping criteria for convergence and feasibility.

Simulations were conducted using the high-fidelity software framework established \hide[by]{in our previous work} \cite{jia2023performance}. The host vehicle is modeled as an electric vehicle prototype tasked with tracking a leading vehicle following a predefined driving cycle. The cycle conforms to the Real Driving Emissions (RDE) requirements under the European regulation~\cite{EC2016_427}.

A total driving duration of approximately 7 minutes was simulated, with the MPC controller executing at a sampling interval of 100~ms. This resulted in approximately 4000 optimization problems solved throughout the scenario. The simulation captures a variety of realistic urban driving conditions, consistent with typical real-world driving patterns.

 
\subsection{Solver Comparison under Different Prediction Horizons}
\noindent 
A comparative performance analysis of the two optimization approaches was carried out across different prediction horizons \( N \in \{3,\; 6,\; 20\} \). While a horizon length of \( N = 3 \) was considered in~\cite{bazzana2024augmented}, our prior experience suggests that horizons in the range of $N = 5, \ldots, 7$ are more appropriate for the MACC application. To evaluate solver performance at larger problem scales, we also include a longer horizon of \( N = 20 \).

The problem complexity scales linearly with the horizon length \( N \): the number of decision variables is \( X = 7N \), with \( l = 2N \) equality constraints and \( q = 13N \) inequality constraints.
Both solvers were configured with a maximum of 300 outer iterations and up to 10 iterations per inner loop. All Lagrange multipliers were initialized to zero. \\For the BIPM solver, the barrier method parameters were set to \( \kappa_0 = 0.5 \), \( \kappa_{\text{final}} = 1500 \), with an update factor \( \nu = 8 \).
For the AL solver, we adopted the algorithm described in~\cite[\S IV.D.2]{abdelkarim2025factor}, initializing the penalty parameters as \( \rho_0 = 0.5 \), \( \rho_{\max} = 5\text{e}5 \), and updating them by a factor \( \rho_{\nu} = 20 \). 
To enhance convergence and ensure feasibility, warm-start initialization was applied to the variable nodes in all cases.
 
Table~\ref{tab:results_performance_comparison} presents a comparative analysis of solver performance for BIPM and AL across varying MPC prediction horizons. The reported metrics include the average, maximum, and minimum number of total iterations, along with the corresponding solver computation times in milliseconds.

The BIPM solver consistently outperforms the AL method in terms of both iteration count and computational time across all tested horizons. For example, at \( N = 6 \), BIPM required an average of only 24.7 iterations compared to 37.9 for AL, while also achieving a faster average computation time (32.1~ms vs. 41.3~ms). The performance gap becomes more pronounced at larger horizons: for \( N = 20 \), BIPM reduced the average number of iterations by approximately 29\%, while the AL reached the maximum iteration limit. These results demonstrate the superior scalability and computational efficiency of the BIPM solver as problem complexity increases.

\begin{table}[htbp]
	\centering
     \scriptsize
	\caption{Performance comparison of AL and BIPM methods}
	\resizebox{8.8cm}{!}{ 
		\begin{tabular}{lclll} 
			\toprule
			& \multirow{2}{*}{$N$} & \multicolumn{3}{c}{\blue{Iterations Number} ; \red{Calculation time (ms)}} \\
			\cmidrule{3-5} 
			& & \multicolumn{1}{c}{avg} & \multicolumn{1}{c}{max} & \multicolumn{1}{c}{min} \\
			\midrule
			
			AL & \multirow{2}{*}{3} 
			& \blue{31.3} ; \red{17.9} 
			& \blue{76} ; \red{76.5} 
			& \blue{\textbf{18}} ; \red{\textbf{10.2}} \\
			
			BIPM & 
			& \blue{\textbf{23.6}} ; \red{\textbf{16.6}} 
			& \blue{\textbf{36}} ; \red{\textbf{62.9}} 
			& \blue{\textbf{18}} ; \red{12.5} \\
			\midrule
			
			AL & \multirow{2}{*}{6} 
			& \blue{37.9} ; \red{41.3} 
			& \blue{89} ; \red{120.9} 
			& \blue{20} ; \red{\textbf{22.5}} \\
			
			BIPM & 
			& \blue{\textbf{24.7}} ; \red{\textbf{32.1}} 
			& \blue{\textbf{65}} ; \red{\textbf{106.5}} 
			& \blue{\textbf{19}} ; \red{24.6} \\
			\midrule
			
			AL & \multirow{2}{*}{20} 
			& \blue{49.5} ; \red{170.7} 
			& \blue{300} ; \red{\textbf{1042.6}} 
			& \blue{31} ; \red{108.5} \\
			
			BIPM & 
			& \blue{\textbf{35.3}} ; \red{\textbf{151.3}} 
			& \blue{\textbf{263}} ; \red{1178.3} 
			& \blue{\textbf{23}} ; \red{\textbf{98.8}} \\
			\bottomrule
		\end{tabular}%
	}
	\label{tab:results_performance_comparison}
\end{table}

\begin{figure}
	\centering
	\includegraphics[trim={0cm 0.9cm 0cm 0cm}, width=.48\textwidth]{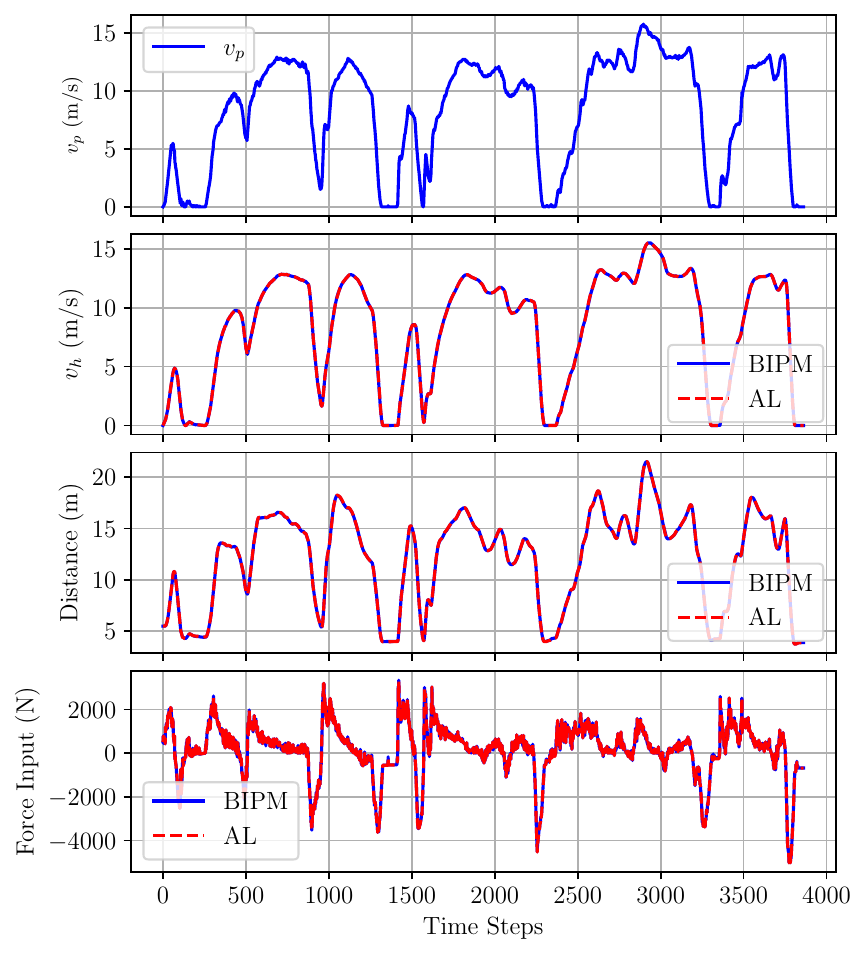}
    \caption{Comparison of control performance between BIPM and AL solvers applied to the MACC problem.}
	\label{fig:solver_comparison}
\end{figure}

Figure~\ref{fig:solver_comparison} illustrates the closed-loop behavior of the MACC problem using both the BIPM and AL solvers under identical configurations. The subplots show the reference and actual velocities of the preceding and following vehicles, the inter-vehicle distance, and the control force input over time. Both solvers produce nearly indistinguishable tracking performance and control effort, which is expected since they operate under the same prediction model, constraints, and controller settings. This confirms that, when appropriately tuned, both optimization methods can achieve comparable control quality despite their differing algorithmic structures. However, as detailed below, parameter tuning tends to be more challenging and sensitive in the case of the AL algorithm.

\subsection{Solver Performance Under Parameter Tuning}
\noindent To evaluate the impact of parameter tuning on solver performance, we present Tables~\ref{tab:parameter_tuning} and~\ref{tab:al_tuning}, which summarize the behavior of the BIPM and AL solvers under various configurations. Cases~2 and~8, highlighted in bold, were previously presented in Table~\ref{tab:results_performance_comparison} for a horizon length of $N = 6$. The parameters are varied based on these two baseline cases in order to examine the influence of parameter tuning on solver performance, considering them as reference configurations. In these tables, {Inner Iter} denotes the maximum number of inner iterations allowed. The standard deviation (SD) reflects the variability in the total number of iterations across multiple runs, with higher values indicating less consistent convergence behavior.

 \begin{table}[t]
 	\centering
 	\caption{Influence of parameter tuning on BIPM solver performance.}
 	\begin{tabular}{cccccccc}
 		\toprule
 		Case & $\nu$ & $\kappa_0$ & $\kappa_{\text{final}}$ & Inner & Avg. & Max & SD \\
 		&  &  &  & Iter & Iter & Iter & \\
 		\midrule
 		1 & 4 & 0.05 & $1.5\cdot10^3$ & 10 & 27.3 & 56 & 2.2 \\
 		\textbf{2} & \textbf{8} & \textbf{0.5} & \textbf{$\textbf{1.5}\cdot\textbf{10}^\textbf{3}$} & \textbf{10} & \textbf{24.7} & \textbf{65} & \textbf{2.5} \\
 		3 & 50 & 0.5 & $1.5\cdot10^3$ & 10 & 24.4 & 155 & 6.9 \\
 		4 & 50 & 0.5 & $1.5\cdot10^3$ & 25 & 29.9 & 58 & 3.2 \\
 		5 & 8 & 10 & $1.5\cdot10^3$ & 10 & 21.2 & 110 & 9.8 \\
 		6 & 8 & 10 & $1.5\cdot10^3$ & 25 & 22.5 & 73 & 4.8 \\
 		7 & 20 & 2 & $1.5\cdot10^6$ & 15 & 27.6 & 70 & 3.2 \\
 		\bottomrule
 	\end{tabular}
 	\label{tab:parameter_tuning}
 \end{table}
 \paragraph{BIPM Solver Tuning Analysis.}
Cases 1–3 demonstrate the impact of the barrier update factor 
$\nu$ on solver performance, aligning with the theoretical framework of Boyd \etal \cite{boyd2004convex}. The barrier parameter update factor $\nu$ exhibits a key trade-off: smaller values (e.g., $\nu=8$) increase outer iterations but require fewer inner steps per iteration, while larger values (e.g., $\nu=50$) reduce outer iterations at the expense of increased inner computation. Notably, the total iteration count remains relatively insensitive to $\nu$ variations. Case 3, with $\nu = 50$ and only 10 inner iterations, shows higher standard deviation, indicating difficulty in following the central path—the trajectory of solutions to the barrier problem as the barrier parameter decreases. Increasing the maximum number of inner iterations in Case 4 stabilizes the solver by allowing closer tracking of this path.

Cases 5--6 highlight the effect of initial barrier value $\kappa_0$. A large $\kappa_0$ increases iteration variance (Case 5), but allowing more inner steps (Case 6) improves stability by better aligning with the central path early on.

Finally, Case 7 demonstrates that increasing the final barrier value $\kappa_{\text{final}}$ by a factor of $10^3$ increases the average iterations only modestly, suggesting the solver is robust to this parameter when others are well tuned.

\begin{table}[t]
	\centering
	\caption{Influence of penalty parameter tuning on AL solver performance.}
	\begin{tabular}{cccccccc}
		\toprule
		Case & $\rho_\nu$ & $\rho_0$ & $\rho_{\max}$ & Inner & Avg. & Max & SD \\
		&  &  &  & Iter & Iter & Iter & \\
		\midrule
		\textbf{8}  & \textbf{20} &\textbf{ 0.5} & $\textbf{5}\cdot\textbf{10}^\textbf{5}$ &\textbf{10}& \textbf{37.9} & \textbf{89}  & \textbf{9.9 }\\
		9  & 5  & 0.5 & $5\cdot10^5$ & 10 & 49.3 & 101 & 8.8 \\
		10 & 50 & 0.5 & $5\cdot10^5$ & 10 & 33.6 & 173 & 10.0 \\
		11 & 50 & 0.5 & $5\cdot10^5$ & 25 & 34.2 & 300 & 11.9 \\
		12 & 20 & 10  & $5\cdot10^5$ & 10 & 32.1 & 157 & 8.4 \\
		13 & 20 & 10  & $5\cdot10^5$ & 25 & 32.1 & 300 & 10.8 \\
		14 & 20 & 0.5 & $5\cdot10^5$ & 25 & 38.6 & 143 & 11.3 \\
		15 & 20 & 0.5 & $5\cdot10^4$ & 10 & 69.8 & 300 & 72.3 \\
		16 & 20 & 0.5 & $5\cdot10^6$ & 10 & 40.1 & 70  & 7.3 \\
		\bottomrule
	\end{tabular}
	\label{tab:al_tuning}
\end{table}

\paragraph{AL Solver Tuning Analysis.}
Cases 8–10 highlight the effect of the penalty update factor $\rho_\nu$. While increasing $\rho_\nu$ reduces the average number of iterations (e.g., Case 10 vs. 9), it also increases variation and the likelihood of hitting the iteration limit, as seen in the spike in maximum iterations in Case 10. In contrast, BIPM (Cases 1, 2, and 4) demonstrates more stable behavior with lower variance and predictable performance in similar scenarios.

Increasing the maximum number of inner iterations in AL (Cases 11, 13, and 14) did not yield improved stability. For instance, Case 11 reaches the iteration cap without significant improvement over Case 10. This contrasts with BIPM, where increasing inner iterations (Case 4 vs. Case 3) helped track the central path more effectively, reducing variance and improving convergence.

The initial penalty value $\rho_0$ also has a complex effect. In Case 12 vs. Case 8, raising $\rho_0$ decreased the average iteration count but led to higher maximum iteration counts—again highlighting AL’s sensitivity to parameter changes.

Finally, adjusting $\rho_{\max}$ has a significant impact on solver behavior. Reducing $\rho_{\max}$ by a factor of 10 (i.e., from $5 \cdot 10^5$ to $5 \cdot 10^4$, as in Case 15) severely degrades performance. Conversely, increasing it to $5 \cdot 10^6$ (Case 16) alters the iteration behavior but does not necessarily improve convergence. In fact, excessively large values of $\rho_{\max}$ may lead to numerical instability and inconsistent progress. For example, in~\cite{bazzana2024augmented}, the penalty parameter is constrained to the range $[0.5,\; 2]$, which proves to be unsuitable for our application, where larger and more adaptive scaling is necessary for acceptable solver performance.

\textbf{In summary}, while BIPM tuning shows predictable and interpretable effects tied to theoretical properties (e.g., central path tracking), AL solver performance is more sensitive and less predictable, often requiring trial-and-error for practical tuning.

\section{Conclusion and Future Work}
\label{sec:conclusion}
\noindent 
This work establishes a principled integration of Barrier Interior Point Methods (BIPM) into factor graph optimization also with inequality constraints as for instance required in optimal control tasks and especially MPC. By introducing inequality factor nodes encoding logarithmic barrier functions, we enable factor graphs to natively support inequality-constrained problems. Our implementation as an extension to the g2o library supports both BIPM and Augmented Lagrangian (AL) methods under a unified frontend, allowing seamless problem formulation with backend-specific solvers.

Empirical results from a multi-objective adaptive cruise control (MACC) task demonstrate the advantages of BIPM in terms of convergence speed and computational efficiency, particularly under large-scale problem settings. The proposed solver achieves comparable control performance to AL while offering improved robustness to hyperparameter tuning and problem scaling.

Looking ahead, this foundational integration opens several promising directions.  From an algorithmic perspective, incorporating more advanced IPM variants (e.g., predictor-corrector or infeasible-start methods) could further accelerate convergence and enhance numerical stability.

\addtolength{\textheight}{-12cm}   


\bibliographystyle{IEEEtran} 
\bibliography{section/10-bibliography}

\begin{thebibliography}{10}
\providecommand{\url}[1]{#1}
\csname url@samestyle\endcsname
\providecommand{\newblock}{\relax}
\providecommand{\bibinfo}[2]{#2}
\providecommand{\BIBentrySTDinterwordspacing}{\spaceskip=0pt\relax}
\providecommand{\BIBentryALTinterwordstretchfactor}{4}
\providecommand{\BIBentryALTinterwordspacing}{\spaceskip=\fontdimen2\font plus
\BIBentryALTinterwordstretchfactor\fontdimen3\font minus
  \fontdimen4\font\relax}
\providecommand{\BIBforeignlanguage}[2]{{%
\expandafter\ifx\csname l@#1\endcsname\relax
\typeout{** WARNING: IEEEtran.bst: No hyphenation pattern has been}%
\typeout{** loaded for the language `#1'. Using the pattern for}%
\typeout{** the default language instead.}%
\else
\language=\csname l@#1\endcsname
\fi
#2}}
\providecommand{\BIBdecl}{\relax}
\BIBdecl

\bibitem{dellaert2017factor}
F.~Dellaert and M.~Kaess, ``Factor graphs for robot perception,''
  \emph{Foundations and Trends{\textregistered} in Robotics}, vol.~6, no. 1-2,
  pp. 1--139, 2017.

\bibitem{bavle2022s}
H.~Bavle, J.~L. Sanchez-Lopez, M.~Shaheer, J.~Civera, and H.~Voos, ``S-graphs+:
  Real-time localization and mapping leveraging hierarchical representations,''
  \emph{IEEE Robotics and Automation Letters}, 2022.

\bibitem{bemporad2006model}
A.~Bemporad, ``Model predictive control design: New trends and tools,'' in
  \emph{Proceedings of the 45th IEEE Conference on Decision and Control}.\hskip
  1em plus 0.5em minus 0.4em\relax IEEE, 2006, pp. 6678--6683.

\bibitem{Abdelkarim2020Development}
A.~Abdelkarim, ``Development of numerical solvers for online optimization with
  application to mpc-based energy-optimal adaptive cruise control,'' master
  thesis, Technische Universit{\"a}t Kaiserslautern, 2020. Available online at
  \url{http://dx.doi.org/10.13140/RG.2.2.11897.28000}, accessed: 2025-05-15.

\bibitem{bazzana2024augmented}
B.~Bazzana, H.~Andreasson, and G.~Grisetti, ``How-to augmented lagrangian on
  factor graphs,'' \emph{IEEE Robotics and Automation Letters}, 2024.

\bibitem{andersson2019casadi}
J.~A. Andersson, J.~Gillis, G.~Horn, J.~B. Rawlings, and M.~Diehl, ``Casadi: a
  software framework for nonlinear optimization and optimal control,''
  \emph{Mathematical Programming Computation}, vol.~11, pp. 1--36, 2019.

\bibitem{abdelkarim2020optimal}
A.~Abdelkarim and P.~Zhang, ``Optimal scheduling of preventive maintenance for
  safety instrumented systems based on mixed-integer programming,'' in
  \emph{Model-Based Safety and Assessment: 7th International Symposium, IMBSA
  2020, Lisbon, Portugal, September 14--16, 2020, Proceedings 7}.\hskip 1em
  plus 0.5em minus 0.4em\relax Springer, 2020, pp. 83--96.

\bibitem{abdelkarim2023optimization}
A.~Abdelkarim, Y.~Jia, and D.~Gorges, ``Optimization of vehicle-to-grid
  profiles for peak shaving in microgrids considering battery health,'' in
  \emph{IECON 2023-49th Annual Conference of the IEEE Industrial Electronics
  Society}.\hskip 1em plus 0.5em minus 0.4em\relax IEEE, 2023, pp. 1--6.

\bibitem{abdelkarim2025factor}
A.~Abdelkarim, H.~Voos, and D.~G{\"o}rges, ``Factor graphs in
  optimization-based robotic control-a tutorial and review,'' \emph{IEEE
  Access}, 2025.

\bibitem{abdelkarim2025ecg2o}
------, ``ecg2o: A seamless extension of g2o for equality-constrained factor
  graph optimization,'' \emph{arXiv preprint arXiv:2503.01311}, 2025.

\bibitem{bertsekas2014constrained}
D.~P. Bertsekas, \emph{Constrained optimization and Lagrange multiplier
  methods}.\hskip 1em plus 0.5em minus 0.4em\relax Academic press, 2014.

\bibitem{boyd2004convex}
S.~P. Boyd and L.~Vandenberghe, \emph{Convex optimization}.\hskip 1em plus
  0.5em minus 0.4em\relax Cambridge university press, 2004.

\bibitem{forsgren2002interior}
A.~Forsgren, P.~E. Gill, and M.~H. Wright, ``Interior methods for nonlinear
  optimization,'' \emph{SIAM review}, vol.~44, no.~4, pp. 525--597, 2002.

\bibitem{bleyer2018advances}
J.~Bleyer, ``Advances in the simulation of viscoplastic fluid flows using
  interior-point methods,'' \emph{Computer Methods in Applied Mechanics and
  Engineering}, vol. 330, pp. 368--394, 2018.

\bibitem{grisetti2010tutorial}
G.~Grisetti, R.~K{\"u}mmerle, C.~Stachniss, and W.~Burgard, ``A tutorial on
  graph-based slam,'' \emph{IEEE Intelligent Transportation Systems Magazine},
  vol.~2, no.~4, pp. 31--43, 2010.

\bibitem{dellaert2021factor}
F.~Dellaert, ``Factor graphs: Exploiting structure in robotics,'' \emph{Annual
  Review of Control, Robotics, and Autonomous Systems}, vol.~4, pp. 141--166,
  2021.

\bibitem{kaess2012isam2}
M.~Kaess, H.~Johannsson, R.~Roberts, V.~Ila, J.~J. Leonard, and F.~Dellaert,
  ``isam2: Incremental smoothing and mapping using the bayes tree,'' \emph{The
  International Journal of Robotics Research}, vol.~31, no.~2, pp. 216--235,
  2012.

\bibitem{chen2019LQR}
G.~Chen and F.~Zhang, Yetongand~Dellaert, ``Lqr control using factor graphs,''
  2019, available online at:
  \url{https://gtsam.org/2019/11/07/lqr-control.html},accessed: 2024-06-15.

\bibitem{yang2021equality}
S.~Yang, G.~Chen, Y.~Zhang, H.~Choset, and F.~Dellaert, ``Equality constrained
  linear optimal control with factor graphs,'' in \emph{2021 IEEE International
  Conference on Robotics and Automation (ICRA)}.\hskip 1em plus 0.5em minus
  0.4em\relax IEEE, 2021, pp. 9717--9723.

\bibitem{chen2022gtgraffiti}
G.~Chen, S.~Baek, J.-D. Florez, W.~Qian, S.-w. Leigh, S.~Hutchinson, and
  F.~Dellaert, ``Gtgraffiti: Spray painting graffiti art from human painting
  motions with a cable driven parallel robot,'' in \emph{2022 International
  Conference on Robotics and Automation (ICRA)}.\hskip 1em plus 0.5em minus
  0.4em\relax IEEE, 2022, pp. 4065--4072.

\bibitem{darnley2021flow}
R.~Darnley, ``Flow control of wireless mesh networks using lqr and factor
  graphs,'' master thesis, Carnegie Mellon University, 2021.

\bibitem{xie2020factor}
M.~Xie, A.~Escontrela, and F.~Dellaert, ``A factor-graph approach for
  optimization problems with dynamics constraints,'' \emph{arXiv preprint
  arXiv:2011.06194}, 2020.

\bibitem{king2022simultaneous}
M.~King-Smith, P.~Tsiotras, and F.~Dellaert, ``Simultaneous control and
  trajectory estimation for collision avoidance of autonomous robotic
  spacecraft systems,'' in \emph{2022 International Conference on Robotics and
  Automation (ICRA)}.\hskip 1em plus 0.5em minus 0.4em\relax IEEE, 2022, pp.
  257--264.

\bibitem{yang2023tightly}
P.~Yang and W.~Wen, ``Tightly joining positioning and control for trustworthy
  unmanned aerial vehicles based on factor graph optimization in urban
  transportation,'' in \emph{2023 IEEE 26th International Conference on
  Intelligent Transportation Systems (ITSC)}.\hskip 1em plus 0.5em minus
  0.4em\relax IEEE, 2023, pp. 3589--3596.

\bibitem{yang2024tightly}
P.~Yang, W.~Wen, S.~Bai, and L.-T. Hsu, ``Tightly joined positioning and
  control model for unmanned aerial vehicles based on factor graph
  optimization,'' \emph{arXiv preprint arXiv:2404.14724}, 2024.

\bibitem{sodhi2020ics}
P.~Sodhi, S.~Choudhury, J.~G. Mangelson, and M.~Kaess, ``Ics: Incremental
  constrained smoothing for state estimation,'' in \emph{2020 IEEE
  International Conference on Robotics and Automation (ICRA)}.\hskip 1em plus
  0.5em minus 0.4em\relax IEEE, 2020, pp. 279--285.

\bibitem{qadri2022incopt}
M.~Qadri, P.~Sodhi, J.~G. Mangelson, F.~Dellaert, and M.~Kaess, ``Incopt:
  Incremental constrained optimization using the bayes tree,'' in \emph{2022
  IEEE/RSJ International Conference on Intelligent Robots and Systems
  (IROS)}.\hskip 1em plus 0.5em minus 0.4em\relax IEEE, 2022, pp. 6381--6388.

\bibitem{steck2018lagrange}
D.~Steck, ``Lagrange multiplier methods for constrained optimization and
  variational problems in banach spaces,'' Ph.D. dissertation, Universit{\"a}t
  W{\"u}rzburg, 2018.

\bibitem{abdelkarim2023accelerated}
A.~Abdelkarim, Y.~Jia, and D.~G{\"o}rges, ``An accelerated interior-point
  method for convex optimization leveraging backtracking mitigation,'' in
  \emph{IECON 2023-49th Annual Conference of the IEEE Industrial Electronics
  Society}.\hskip 1em plus 0.5em minus 0.4em\relax IEEE, 2023, pp. 1--6.

\bibitem{jia2023performance}
Y.~Jia, A.~Abdelkarim, X.~Klingbeil, R.~Savelsberg, and D.~G{\"o}rges,
  ``Performance evaluation of energy-optimal adaptive cruise control in
  simulation and on a test track,'' \emph{IFAC-PapersOnLine}, vol.~56, no.~2,
  pp. 4994--5000, 2023.

\bibitem{EC2016_427}
\BIBentryALTinterwordspacing
{European Commission}, ``Commission regulation (eu) no 2016/427,''
  \emph{Official Journal of the European Union}, vol. L 82, pp. 1--98, 2016.
  [Online]. Available: \url{https://eur-lex.europa.eu/eli/reg/2016/427/oj}
\BIBentrySTDinterwordspacing

\end{thebibliography}

\end{document}